\documentclass[letterpaper, 10 pt, conference]{ieeeconf}  

\IEEEoverridecommandlockouts                              

\usepackage{graphicx}
\graphicspath{{figs/}}
\usepackage{amsmath}
\usepackage{amssymb}
\usepackage{algorithm}
\usepackage{algpseudocode}
\usepackage{times}
\usepackage{bm}
\usepackage{color}
\usepackage{mathtools}
\usepackage{algorithm}
\usepackage{algpseudocode}
\usepackage{multirow}
\usepackage{hyperref}
\newcommand{\figref}[1]{Fig.~\ref{#1}}
\newcommand{\tabref}[1]{Table~\ref{#1}}
\newcommand{\secref}[1]{Section~\ref{#1}}
\newcommand{\algoref}[1]{Algorithm~\ref{#1}}
\newcommand{\R}{\mathbb{R}}

\newcommand{\SO}{\mathop{\mathrm{SO}}}
\newcommand{\Ortho}{{\mathrm{O}}}

\newcommand{\erfc}{\mathop{\mathrm{erfc}}}

\newcommand{\TODO}[1]{\textcolor{red}{\textbf{TODO}: #1}}
\renewcommand{\TODO}[1]{\relax}

\newcommand{\D}{{D}}
\newcommand{\eigvals}{\bm{\lambda}}
\newcommand{\normconst}{\mathcal{C}}
\newcommand{\integrand}{\mathcal{F}}
\newcommand{\triu}{\mathop{\mathrm{triu}}}
\newcommand{\diag}{\mathop{\mathrm{diag}}}

\newcommand{\OmegaL}{\mathop{\mathrm{\Omega_L}}}
\newcommand{\OmegaR}{\mathop{\mathrm{\Omega_R}}}
\newcommand{\qt}{\bm{q}}
\newcommand{\conj}[1]{{#1^*}}
\newcommand{\qtc}{\conj{\bm{q}}}
\newcommand{\Sp}{\mathbb{S}}
\newcommand{\bingham}{\mathfrak{B}}
\newcommand{\Sym}{\mathrm{Sym}}

\newcommand{\etal}{\textit{et al.}~}

\renewcommand{\d}{\mathrm{d}}

\title{\LARGE \bf
A Probabilistic Rotation Representation for Symmetric Shapes With an Efficiently Computable Bingham Loss Function$^\text{*}$
}

\author{Hiroya Sato$^\text{1}$, Takuya Ikeda$^\text{2}$, and Koichi Nishiwaki$^{2}$
\thanks{This work has been submitted to the IEEE for possible publication. Copyright may be transferred without notice, after which this version may no longer be accessible.}%
\thanks{$^{\text{*}}$ This work has been done at Woven Planet Holdings, Inc.
A part of the work is the result of Summer Internship Program.}%
\thanks{$^{\text{1}}$Hiroya Sato is with Department of Mechano-Informatics, Graduate School
of Information Science and Technology, The University of Tokyo, 7-3-1
Hongo, Bunkyo-ku, Tokyo, 113-8656, Japan.
{\tt\footnotesize h-sato@jsk.t.u-tokyo.ac.jp}
}%
\thanks{$^{\text{2}}$ Authors are with the Woven Planet Holdings, Inc. 3 Chome-2-1 Nihonbashimuromachi, Chuo City, Tokyo, 103-0022, Japan, {\tt\footnotesize [firstname.lastname]@woven-planet.global}}%
}%

\begin{document}

\maketitle
\thispagestyle{empty}
\pagestyle{empty}

\begin{abstract}

In recent years, a deep learning framework has been widely used for object pose estimation. 
While quaternion is a common choice for rotation representation, it cannot represent the ambiguity of the observation.
In order to handle the ambiguity, the Bingham distribution is one promising solution.
However, it requires complicated calculation when yielding the negative log-likelihood (NLL) loss.
An alternative easy-to-implement loss function has been proposed to avoid complex computations but has difficulty expressing symmetric distribution.
In this paper, we introduce a fast-computable and easy-to-implement NLL loss function for Bingham distribution. 
We also create the inference network and show that our loss function can capture the symmetric property of target objects from their point clouds.

\end{abstract}

\section{INTRODUCTION}

Recently, many research efforts have been on pose estimation based on a deep learning framework, such as \cite{xiang2018posecnn, Bui2018-mr}. In these works, the quaternion is widely used for rotation representation. However, since a single quaternion can only represent a single rotation, it cannot capture the observation's uncertainty. Handling uncertainty is quite important, especially when a target object is occluded or has a symmetric shape \cite{manhardt2019explaining, Hashimoto2019}.

Many researchers have been considering how to represent the ambiguity of rotations. 
In recent years, non-parametric probabilistic representations
over $\SO(3)$, the space of the spatial rotation, 
using neural networks have been developed \cite{implicitpdf2021, epropnp}.
On the other hand, when comparing distributions estimated by other methods that do not use neural networks, or when analyzing distributions with mathematical methods, it is convenient to use explicitly parameterized distributions.
One way to directly parameterize the distribution over $\SO(3)$
is to utilize \textit{Bingham distribution} \cite{bingham1974}. It mainly has two advantages. Firstly, 
the Bingham distribution
(details described in \secref{section:bingham}) 
is easy to parameterize. Secondly, the continuous representation, suitable for a neural network, can be derived from this distribution \cite{peretroukhin_so3_2020}. For the above characteristics, we choose the Bingham distribution for probabilistic rotation representation. 

To optimize the probability distribution, in general, a negative log-likelihood (NLL) is a common choice for loss function. NLL is desirable because it can directly estimate the distribution parameter from given data without additional assumptions. However, the Bingham distribution has difficulty calculating its normalizing constants, which is required in the calculation of NLL and needs to recalculate when the distribution parameter changes. Because this calculation is too complicated to calculate at every iteration, to our best knowledge, a pre-computed table of the constants has been needed. To avoid the complex calculation, Peretroukhin \etal \cite{peretroukhin_so3_2020} suggested a new loss function based on the QCQP method, giving up estimating the full distribution information. 

\begin{figure}[t]
    \centering
    \includegraphics[width=\linewidth]{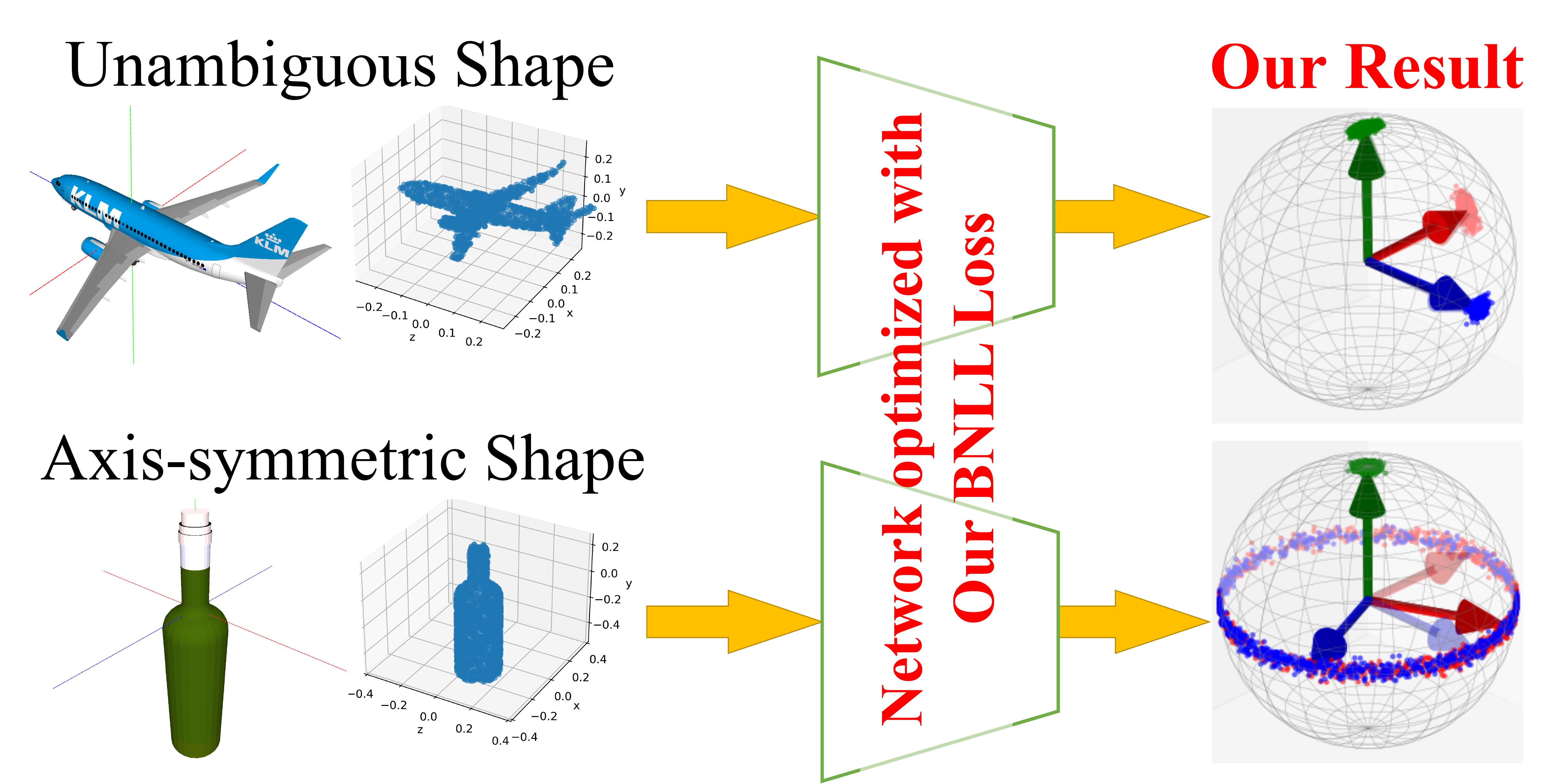}
    \caption{Inference sample of our network. 
    A translucent frame represents the mode of ground truth and an opaque frame the estimated mode.
    One can see that the inference result for point cloud of unambiguous shape (airplane) is unimodal and peaky, while that of axis-symmetric shape (wine bottle) becomes zonally spread around the axis.}
    \vspace{-0.25cm}
    \label{fig:mugbingham}
\end{figure}

In this paper, we introduce a fast-computable and easy-to-implement Bingham NLL (BNLL) loss function, enabling us to introduce BNLL loss with as much effort as the QCQP loss. In addition, we show that BNLL captures the axis-symmetric property of the target distribution, while it is difficult for QCQP to capture the symmetry.  

%

\section{RELATED WORKS}

\subsection{Continuous Rotation Representations}
\label{section:rotationrepresentations}

4-dimensional rotation representation is widely used.
In particular, the quaternion is a popular representation.
It is utilized such as in PoseCNN \cite{xiang2018posecnn}, PoseNet \cite{posenet2015}, and 6D-VNet \cite{Wenbin2021}.
Another 4-dimensional representation is an axis-rotation representation, introduced such as in MapNet \cite{mapnet2018}.
Although these 4D representation are valid in some cases, it is known that every $d$-dimensional ($d<5$) rotation representation is ``discontinuous'' (in the sense of \cite{Zhou2019}), since the 3-dimensional real projective space $\R P^3 (\cong \SO(3))$
cannot be embedded in $\R^d$ unless $d \geq 5$ \cite{Davis1998EmbeddingsOR}.
Because these representations have some singular points that prevent the network from stable regression,
it is preferred that we use the continuous rotation representation in training neural networks.

Researchers have proposed
various high-dimension representations.
For example, 9D representation was proposed in \cite{levinson20neurips} by using the singular value decomposition (SVD). 10D representation was proposed in \cite{peretroukhin_so3_2020} using the 4-dimensional symmetric matrix, which can be used for the parametrization of Bingham distribution.
We will examine other continuous paramterization in \secref{section:rotationalambiguity}.

\subsection{Expressions of Rotational Ambiguity}
\label{section:rotationalambiguity}

There are several ways to express rotational ambiguity.
In \cite{manhardt2019explaining, Bui2020}, they tried capturing the ambiguity by using multiple quaternions and minimizing the special loss function.
In KOSNet \cite{Hashimoto2019}, they used three parameters for describing the camera's rotation (elevation, azimuth, and rotation around the optical axis) and employed Gaussian distribution for describing their ambiguity.
These two representations, however, are discontinuous as described in \secref{section:rotationrepresentations}, since their dimensions are both lesser than 5.

Another approach is to use Bingham distribution.
This distribution is easy to parameterize and has been utilized in the pose estimation field.
It is used for the model distribution of a Bayesian filter to realize online pose estimation \cite{Srivatsan2017},
for visual self-localization \cite{deng2020deep},
for multiview fusion \cite{Riedel2016},
and for describing the pose ambiguity of objects with symmetric shape \cite{Gilitschenski2020}.
Moreover, 
there is a continuous parameterization of this distribution, 
as Peretroukhin \etal \cite{peretroukhin_so3_2020} gives an example of it.
We adopt Bingham distribution as our probabilistic rotation representation.

\subsection{Loss Functions for Bingham Representation}

A negative log-likelihood (NLL) loss function is a common choice and has advantages, as described in the Introduction.
When computing Bingham NLL (BNLL) loss,
the main barrier is the computation of the normalizing constant.
One solution to this problem is to take the time to pre-compute the table of normalizing constants. For example, Kume \etal \cite{Kume2013saddlepoint} uses the saddlepoint approximation technique to construct this table.
Moreover, we also need to implement a smooth interpolation function as described in \cite{Gilitschenski2020} to compute a constant missing in the table and a derivative of each constant for backpropagation. 

Instead of using the Bingham NLL loss, 
Peretroukhin \etal proposed the method of estimating the Bingham parameter by using the quadratically-constrained quadratic program (QCQP) \cite{peretroukhin_so3_2020}.
In this method, they solve the following equation.
\begin{equation}
    \qt_\text{amax}(A) = \arg \max_{\qt\in \Sp^3} \qt^\top A \qt.
\end{equation}
This is equivalent to eigendecompose $A$ and finding the eigenvector corresponding to the maximum eigenvalue. Using this function, a loss function for QCQP $\mathcal{L}_\text{QCQP}$ is defined below.
\begin{equation}
    \mathcal{L}_\text{QCQP} (A, \qt_\text{gt}) = d_F\left(\qt_\text{amax}(A), \qt_\text{gt}\right)^2 \label{eq:qcqpdef}
\end{equation}
In this paper, we compare the QCQP and our BNLL results and show that ours can 
achieve the performance that overcomes QCQP with the equivalence cost of implementation.

\section{ROTATION REPRESENTATIONS}

\subsection{Quaternion and Spatial Rotation}

\subsubsection{Quaternion}
We introduce symbols $i,j,k$, which satisfies the property:
\begin{equation}
    i^2=j^2=k^2=ijk=-1.
    \label{eq:imaginaryunits}
\end{equation}
A quaternion is an expression of the form:
\begin{equation}
    q = w + xi+yj+zk
    \label{eq:quaternion}
\end{equation}
where $w,x,y,z$ are real numbers. $i,j,k$ are called the imaginary units of the quaternion.
The set of quaternions forms a 4D vector space whose basis is $\{1,i,j,k\}$.
Therefore, we identify a quaternion $q$ defined in \eqref{eq:quaternion} with
\begin{equation}
    \qt = (w,x,y,z)^\top \in \R^4.
\end{equation}

\subsubsection{Product of Quaternions}
For any quaternion $q' = a + bi + cj + dk$, 
we can define a product of quaternions $q'q$ thanks to the rule \eqref{eq:imaginaryunits}.
The set of quaternions forms a \textit{group} by this multiplication.
$q'q$ can also be identified with an element of $\R^4$. We denote it $\qt'\odot \qt \in \R^4$. 
Note that $q'q \neq qq'$ in general. Since $\qt' \odot \qt$ is bilinear w.r.t. $\qt' $ and $ \qt$, we can define matrices $\OmegaL(\qt')$ and $\OmegaR(\qt)$ satisfying
\begin{equation}
    \qt' \odot \qt = \OmegaL(\qt') \qt = \OmegaR(\qt) \qt'.
\end{equation}
$\OmegaL(\qt')$ and $\OmegaR(\qt)$ can be written in closed form:
\begin{minipage}{0.48\linewidth}
\begin{equation}
    \OmegaL(\qt')\! =\!\! 
    \left(\!\!\begin{array}{c@{\hspace{2mm}}c@{\hspace{2mm}}c@{\hspace{2mm}}c}
        a & -b & -c & -d \\
        b & a & -d & c \\
        c & d & a & -b \\
        d & -c & b & a
    \end{array}\!\!
    \right)\!\!, \label{eq:omegal}
    \vspace{2mm}
\end{equation}
\end{minipage}
\begin{minipage}{0.48\linewidth}
\begin{equation}
    \OmegaR(\qt)\! = \!\!\left(\!\!\begin{array}{c@{\hspace{2mm}}c@{\hspace{2mm}}c@{\hspace{2mm}}c}
        w & -x & -y & -z \\
        x & w & z & -y \\
        y & -z & w & x \\
        z & y & -x & w
    \end{array}\!\!
    \right)\!\!. \label{eq:omegar}
    \vspace{2mm}
\end{equation}
\end{minipage}

\subsubsection{Conjugate, Norm, and Unit Quaternion}
The \textit{conjugate} of $\qt$ is defined by $\qtc = (w,-x,-y,-z)^\top$. 
In general, $\conj{(\qt \odot \qt')} = {\qt'}^* \odot \qtc$. 
In particular, if $\qt = \qt'$, then 
\begin{equation}
    \qt \odot \qtc = \qtc \odot \qt = w^2 + x^2 + y^2 + z^2.
\end{equation}
By definition \eqref{eq:omegal} and \eqref{eq:omegar}, we get
\begin{equation}
    \OmegaL(\qtc) = \OmegaL(\qt)^\top, \quad \OmegaR(\qtc) = \OmegaR(\qt)^\top.
    \label{eq:conjandtranspose}
\end{equation}

We define the norm of quaternion $\|\bm{q}\|$ as
\begin{equation}
    \|\bm{q}\| = \sqrt{\qt \odot \qtc} = \sqrt{\qtc \odot \qt}.
\end{equation}
We call $\bm{q}$ a \textit{unit quaternion} if $\|\bm{q}\| = 1$. For a unit quaternion, its inverse coincides with its conjugate: $ \qt^{-1} = \qtc$.
Using \eqref{eq:conjandtranspose}, we can see that $\OmegaL(\qt)$ and $\OmegaR(\qt)$ are both orthogonal:
\begin{equation}
    \OmegaL(\qt)^\top \OmegaL(\qt) = \OmegaR(\qt)^\top \OmegaR(\qt) = I_{4}
\end{equation}
where $\qt$ is any unit quaternion, and $I_{4}$ is the 4-dimensional identity matrix.

We denote the set of unit quaternions $\Sp^3$ because it is homeomorphic to a 3-sphere $\Sp^3$.

\subsubsection{Unit Quaternion and Spatial rotation}
It is well known that unit quaternions can represent spatial rotation. 
A mapping $R: \Sp^3 \to \SO(3)$ defined below is in fact, a group homomorphism:
\begin{align}
    R(\qt) = \left(\hspace{-1.5mm}
    \begin{array}{ccc}
        1-2y^{2}-2z^{2} & -2 w z+2 x y & 2 w y+2 x z \\
        2 w z+2 x y & 1-2x^{2}-2z^{2} & -2 w x+2 y z \\
        -2 w y+2 x z & 2 w x+2 y z & 1-2x^{2}-2y^{2}
    \end{array}\hspace{-1.5mm}\right)\hspace{-1mm}.
    \label{eq:quaternion2so3}
\end{align}
Crucially, antipodal unit quaternions represent the same rotation; namely, $R(-\qt) = R(\qt)$.

\subsubsection{Distance Function of Quaternions}
Using $R(\cdot)$ defined in \eqref{eq:quaternion2so3} and the Frobenius norm $\|\cdot\|_F$, we can define some distance function over $\Sp^3$.
\begin{align}
    d_G(\qt,\qt') &= 2\arccos \left( \left|{\qt^\top \qt'}\right| \right), \label{eq:dist_geo}\\
    \mathop{d_F}(\qt, \qt') &= \left\|R(\qt)-R\left(\qt'\right)\right\|_{F}. \label{eq:dist_frob}
\end{align}
First $d_G$ is the geodesic distance over $\Sp^3$. 
Second, $d_F$ is the Frobenius distance between the rotation matrices.

\subsection{Definition of Bingham Distribution and Its Properties}
\label{section:bingham}

The Bingham distribution \cite{bingham1974} is a probability distribution on the unit sphere $\Sp^{d-1} \subset \R^{d}$ with the property of antipodal symmetry, which is consistent with the quaternion's property. 
We set $d=4$ throughout of this paper because we only consider $\Sp^3$.
We define \textit{Bingham distribution} as follows.
\begin{equation}
    \bingham(A)(\qt) = \frac{1}{\normconst(\eigvals)} \exp\left( \qt^\top A \qt \right),
\end{equation}
where $\qt \in \Sp^3$ and $A \in \Sym_4$.
Here $\Sym_n$ denotes the set of $n$-dimensional symmetric matrices.
Instead of specifying a symmetric matrix, since every symmetric matrices can be diagonalized by some orthogonal matrix, we can determine the distribution as follows:
\begin{equation}
    \bingham(\D, \eigvals)(\qt) = \frac{1}{\normconst(\eigvals)} \exp\left( \qt^\top \D \diag(\eigvals) \D^\top \qt \right), \label{eq:binghamdefinition}
\end{equation}
where $\qt \in \Sp^3, \,\D\in \Ortho(4),$ and $\eigvals \in \R^4$.
Here $\Ortho(n)$ denotes the $n$-dimensional orthogonal group. 
%
Here we define $\diag : \R^m \to \R^{m\times m}$ as below.
\begin{equation}
    \diag : \begin{pmatrix}
        v_1 \\ \vdots \\ v_m
    \end{pmatrix}
    \mapsto \begin{pmatrix}
        v_1 & & \\
         & \ddots & \\
         & & v_m
    \end{pmatrix}
\end{equation}
The factor $\normconst(\eigvals)$ is called a \textit{normalizing constant} of a Bingham distribution $\bingham(\qt; \D, \eigvals)$.
$\normconst(\eigvals)$ is defined as below:
\begin{equation}
    \normconst(\eigvals) = \int_{\qt \in \Sp^3} \exp\left( \qt^\top \diag(\eigvals) \qt \right) \d_{\Sp^3} (\qt)
\end{equation}
where $\d_{\Sp^3}(\cdot)$ is the uniform measure on the $\Sp^3$.
Note that a normalizing constant depends only on $\eigvals$. 

It is easy to check that for any $c \in \R$, 
\begin{equation}
    \bingham(D, \eigvals + c) = \bingham(D, \eigvals)
\end{equation}
where $\eigvals + c = (\lambda_1 + c, \dots, \lambda_4 + c)$. Therefore, we can set $\eigvals$ satisfying
\begin{equation}
    0 = \lambda_1 \geq \lambda_2 \geq \lambda_3 \geq \lambda_4 \label{eq:sortedlambda}
\end{equation}
by sorting a column of $D$ if necessary.
A processed $D$ and a processed $\eigvals$ are denoted as $D_\text{shifted}$, $\eigvals_\text{shifted}$ respectively.
It follows directly from the Rayleigh's quotient formula that
\begin{equation}
    \arg \max_{\qt \in \Sp^3} \bingham(D, \eigvals)(\qt) = \qt_{\lambda_1}
    \label{eq:modequaternion}
\end{equation}
where $\qt_{\lambda_1}$ is a column vector of $D$ corresponding to the maximum entry of $\eigvals$. If we sort $\eigvals$ as \eqref{eq:sortedlambda}, $\qt_{\lambda_1}$ coincides with the left-most column vector of $D$.
If the eigenvalues of $A$ is sorted and shifted so as to satisfy \eqref{eq:sortedlambda}, then we call it $A_\text{shifted}$.
We assume that all parameters are shifted.


To parametrize the Bingham distribution,
we introduce here the 10D parameterization using a symmetric matrix which proposed by Peretroukhin \etal \cite{peretroukhin_so3_2020}.
\begin{equation}
    \triu : \begin{pmatrix}
        \theta_1 \\ \vdots \\ \theta_{10}
    \end{pmatrix}
    \mapsto 
    \begin{pmatrix}
        \theta_{1} & \theta_{2} & \theta_{3} & \theta_{4} \\
        \theta_{2} & \theta_{5} & \theta_{6} & \theta_{7} \\
        \theta_{3} & \theta_{6} & \theta_{8} & \theta_{9} \\
        \theta_{4} & \theta_{7} & \theta_{9} & \theta_{10}
\end{pmatrix}.
\label{eq:10dparam}
\end{equation}
We define the parameterization as $ \R^{10} \ni \bm{\theta} \mapsto  \bingham(\triu(\bm{\theta}) )$.

\begin{algorithm}[tb]
   \caption{Our implementation of the loss function}
   \label{algo:lossfunction}
   \begin{algorithmic}[1]
   
   \Function {Integrator}{$f_\text{integrant}$, $\eigvals$}
       \State $N_\text{min} \gets 15$; $N \gets 200$
       \State $r\gets 2.5$; $\omega_d \gets 0.5$
       \State Define $c$ as in \eqref{eq:vars}; $d \gets c / 2$
       \State Define $h,p_1,p_2$ as in \eqref{eq:vars}
       \State $S\gets 0$
        \For {$n = -N-1,\dots,N$}
        \State $S \gets S + w(|nh|)\cdot f_\text{integrant}(nh, \eigvals)\cdot e^{nh\sqrt{-1}}$ 
        \Statex \Comment{$w$ is defined in \eqref{eq:weightfunc}}
        \EndFor
       \State \Return the real part of $\pi e^c h S$
   \EndFunction
   \State
   \Function {BinghamLoss}{$D$, $\eigvals$, $\qt_\text{gt}$}
        \State $D_\text{shifted}$, $\eigvals_\text{shifted}$ $\gets$ \Call{Sort\&Shift}{$D$, $\eigvals$}
        \State $A_\text{shifted}$ $\gets$ $D_\text{shifted} \diag(\eigvals_\text{shifted}) D_\text{shifted}^\top$
        \State $\normconst \gets $ \Call{Integrator}{$\integrand$, $\eigvals$} \Comment{see \eqref{eq:def_C}}
        \State \Return $-\qt_\text{gt}^\top A_\text{shifted}\, \qt_\text{gt} + \ln \normconst$
    \EndFunction

   \end{algorithmic}
\end{algorithm}

\section{EFFICIENT COMPUTATION OF BNLL LOSS}
\label{section:computationofloss}
\subsection{Definition of BNLL Loss Function}
The negative log-likelihood function (NLL) of the Bingham distribution can be written as follows:
\begin{equation}
    \mathcal{L}_\text{BNLL}(A, \qt_\text{gt}) =-\boldsymbol{q}_{\text{gt}}^{\top} A \boldsymbol{q}_{\text{gt}}+\ln \mathcal{C}(\eigvals)
    \label{eq:defBNLL}
\end{equation}
It had been a hard problem to compute $\normconst(\eigvals)$ until a highly efficient computation method was proposed by \cite{ChenTanaka2021}, on which our implementation of the loss function is mainly based.

\subsection{Calculation of Normalizing Constant and Its Derivative}

The whole procedure is shown in \algoref{algo:lossfunction}.
The parameterization in this section is mainly derived from \cite{ChenTanaka2021}.
Let $r, \omega_d$ be arbitrary real numbers satisfying 
\begin{equation}
    r\geq 2 \quad\text{and}\quad \frac{1}{r} \leq \omega_d \leq 1.
\end{equation}
We chose $r = 2.5,\,\omega_d = 0.5$ here.
Let $c,h,p_1,p_2$ be defined as
\begin{equation}
    \begin{array}{c}
    \displaystyle
    c = \frac{N_\text{min} \pi }{r^2(1+r) \omega_d},\quad h = \sqrt{\frac{2\pi d (1+r)} {\omega_d N}}, \\[12pt]
    \displaystyle
    p_1 = \sqrt{\frac{Nh}{\omega_d}},\quad p_2 = \sqrt{\frac{\omega_d N h}{4}},
    \end{array}
    \label{eq:vars}
\end{equation}
where $d$ is any positive number satisfying $d < c$. We chose $d = c/2$ here.
$N$ is a positive integer satisfying $N \geq N_\text{min}$.
One can choose $N_\text{min}$ arbitrarily; however, a too small $N_\text{min}$ may lead to unstable computation. 
We chose $N_\text{min} = 15$ here.

We define a function $w$ parametrized by $p_1,\,p_2$ in \eqref{eq:vars} as below.
\begin{equation}
    w(x) = \frac{1}{2} \erfc \left( \frac{x}{p_1} - p_2 \right),
    \label{eq:weightfunc}
\end{equation}
where $\erfc$ is the complementary error function:
\begin{equation}
    \erfc(x) = 1-\frac{2}{\sqrt{\pi}} \int_{0}^{x} e^{-t^{2}} d t.
\end{equation}

By setting
\begin{align}
    \integrand (t, \eigvals) &= \prod^4_{k=1} \left(-\lambda_k + t\sqrt{-1} + c\right)^{-1/2}, \\
    \frac{\partial \integrand}{\partial \lambda_i}(t, \eigvals) &= \frac{1}{2}\left( -\lambda_i + t\sqrt{-1} + c \right)^{-1}\integrand(t, \eigvals),
\end{align}\\
for each $i = 1,\dots, 4$, now we can calculate the normalizing constant $\normconst$ as below
\begin{align}
    \normconst(\eigvals) &= \pi e^c h \sum_{n=-N-1}^N w(|nh|)\, \integrand(nh, \eigvals)\, e^{nh\sqrt{-1}}, \label{eq:def_C}\\
    \frac{\partial \normconst}{\partial \lambda_i}(\eigvals) &= \pi e^c h \sum_{n=-N-1}^N w(|nh|)\, \frac{\partial \integrand}{\partial \lambda_i}(nh, \eigvals)\, e^{nh\sqrt{-1}}, \label{eq:def_dCdLam}
\end{align}
for each $i = 1,\dots, 4$.
Although a calculation result of $\normconst(\eigvals)$ and ${\partial \normconst}/{\partial \lambda_i}(\eigvals)$ should exactly be a real number, one may get a complex number with the very small imaginary part.
In our implementation shown in \algoref{algo:lossfunction}, we ignore the imaginary part, assuming that it is sufficiently small.

It is noteworthy that if we set the true value
of the normalizing constant
as $\normconst_\text{truth}(\eigvals)$, we get
\begin{equation}
    |\normconst_\text{truth}(\eigvals) - \normconst(\eigvals)| = O\left(\sqrt{N} e^{-c\sqrt{N}}\right)
\end{equation}
for a constant $c>0$ independent from $N$ \cite{TANAKA201473}.
Here $O$ denotes the Landau's order symbol.
This means that we can achieve any accuracy if we set a large enough $N$.
In this paper, we set $N = 200$ in consideration of the computation time.

\section{EXPERIMENTS}

\begin{table}
    \caption{Values of $A$ and $\eigvals$ in this section}
    \begin{tabular}{c@{\;:\;}c|c@{\;:\;}c}
        $A_\text{init}$ & 
        {\tiny $\left[
        \begin{array}{@{\hspace{3px}}r@{\hspace{5px}}r@{\hspace{5px}}r@{\hspace{5px}}r@{\hspace{3px}}}
            95.69 & 13.72 & 28.38 & 60.61 \\
            & 94.42 & 85.27 & 0.23 \\
            & & 52.12 & 55.20 \\
            & & & 48.54
        \end{array}
        \right]$}
        &
        $\eigvals_\text{init}$ &
        {\tiny $ \left[
        \begin{array}{@{\hspace{3px}}r@{\hspace{3px}}}
            -236.48 \\ -173.72 \\ -85.51 \\ 0.00
        \end{array}
        
        \right]$} \\[15px]\hline
        \rule{0pt}{20px}
        $A_\text{true}$ & 
        \tiny $\left[
        \begin{array}{@{\hspace{3px}}r@{\hspace{5px}}r@{\hspace{5px}}r@{\hspace{5px}}r@{\hspace{3px}}}
            -116.55 & 40.70 & 119.55 & 225.97 \\
            & -147.05 & 145.26 & -280.25 \\
            & & -386.19 & 52.06 \\
            & & & -743.89
        \end{array}
        \right]$
        &
        $\eigvals_\text{true}$ &
        \tiny $\left[
        \begin{array}{@{\hspace{3px}}r@{\hspace{3px}}}
            -926.44 \\ -467.07 \\ -0.17 \\ 0.00
        \end{array}
        \right]$
    \end{tabular}
    \label{tab:values_of_As}
\end{table}

\begin{figure}[t]
    \centering
    \begin{tabular}{@{\hspace{-0.5mm}}c@{\hspace{1mm}}c}
    \begin{minipage}{0.5\linewidth}
        \centering
        \includegraphics[width=\linewidth]{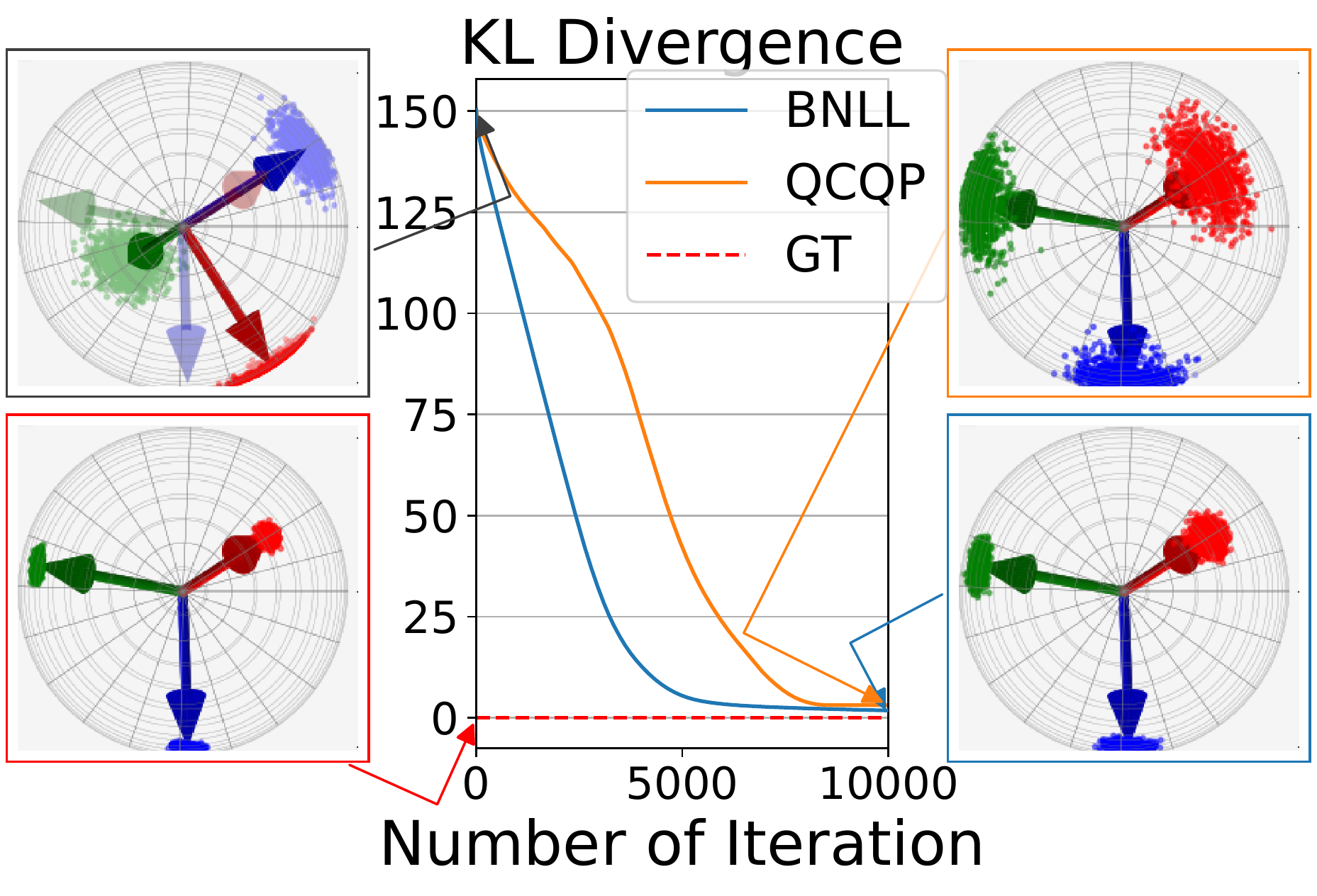}
        \small{(a) Unimodal case}
    \end{minipage}&
    \begin{minipage}{0.5\linewidth}
        \centering
        \includegraphics[width=\linewidth]{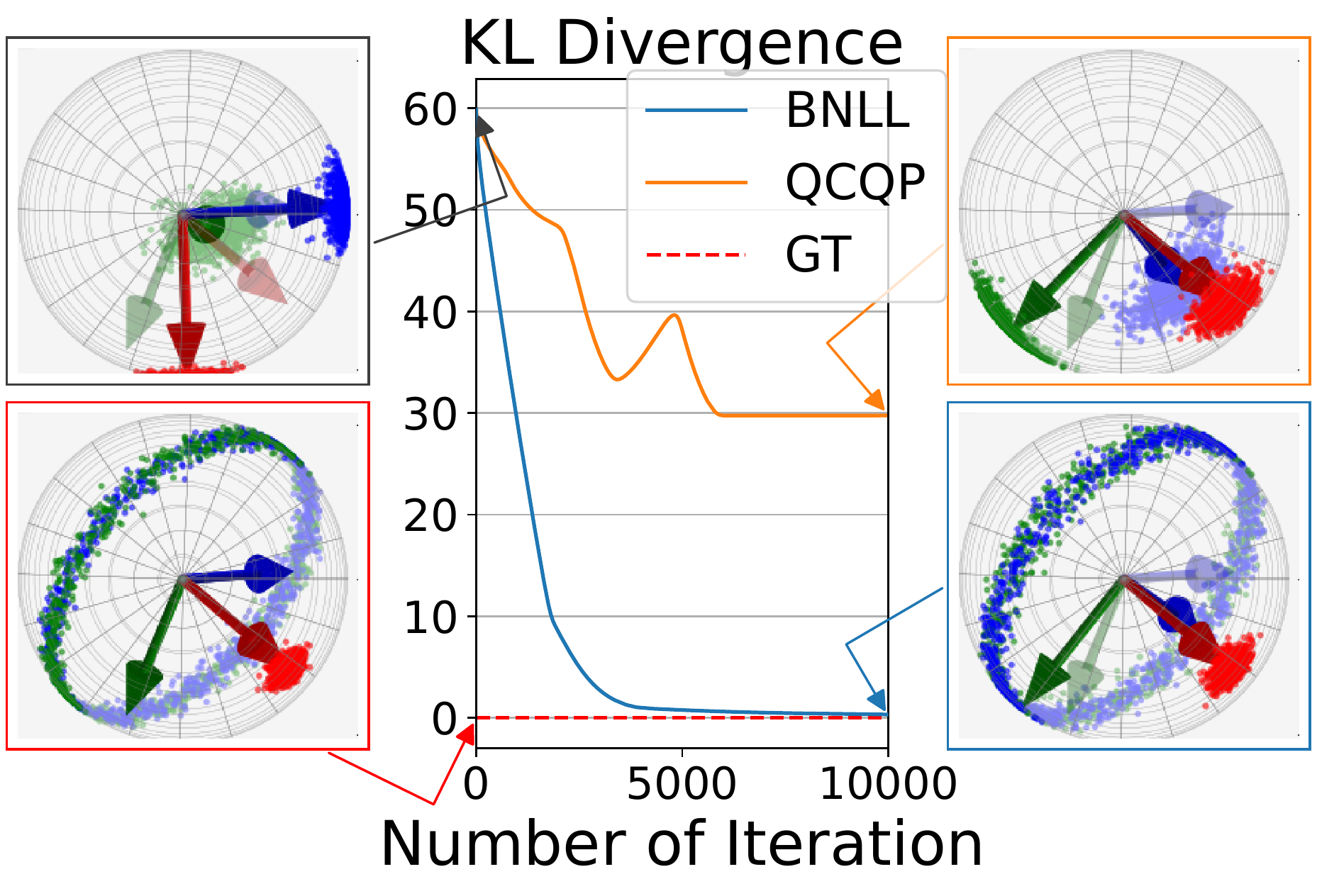}
        \small{(b) Axis-symmetric case}
    \end{minipage}
\end{tabular}
\caption{
The transition of Kullback-Leibler divergence (KLD) during optimization. ``GT'' stands for ``ground truth''.
Each visualization corresponds to the distribution at the tip of the arrow.
The left-top figure is the initial distribution. $N_\text{sample} = 100$ is applied here. The ground truth is $\bingham(A_\text{true})$ shown in \tabref{tab:values_of_As}.
The eigenvalues of true parameter used for unimodal case is $[0, -1209.9, -2217.9, -2342.4]$. $A_\text{init}$ is common.
}
\label{fig:transition_of_KLD}
\end{figure}

\subsection{Experiments with Points Sampled from Distribution}
\label{section:sampledistr}

\subsubsection{Experiments Settings}
To examine the fundamental perfomance of QCQP and BNLL,
we optimized a parameter of Bingham distribution with points pre-sampled from a given Bingham distribution in advance of optimization.
The method in \cite{Kent2013BinghamSampling} is employed in sampling.
We sampled $N_\text{sample}$ points from a distribution $\bingham(A)$ with an arbitrary $A\in \Sym_4$.
In this section, unless otherwise noted, 
$A_\text{init}$ and $A_\text{true}$ shown in \tabref{tab:values_of_As} are used for the parameter of initial and ground truth, respectively. $\eigvals$ is the eigenvalue of $A$ with the corresponding subscript. One can choose any $A$ since we just limited it only for simplicity of explanation.

\subsubsection{Optimization Result of Unimodal Case}
The optimization results are shown in \figref{fig:transition_of_KLD}.
$N_\text{iter}$ stands for the number of iterations and we set its maximum to 20000.
In the unimodal case, we can see that both QCQP and BNLL estimate the distribuion well.
The angular diffence between the estimated and true mode quaternion are 0.11 [deg] for QCQP and 0.15 [deg] for BNLL.
The estimated mode quaternion using QCQP is very accurate, though the dispersion is broader than that of BNLL. 
The resulting Kullback-Leibler divergence (KLD) is 3.127734 for QCQP and 0.700774 for BNLL.

\subsubsection{Optimization Result of Axis-Symmetric Case}

For an axis-symmetric case, 
however, the resulting KLD of QCQP converges to large value (29.802812), while BNLL approaches close to zero (0.133398).
\figref{fig:transition_of_KLD} (b) shows the result of QCQP becomes unimodal, while BNLL captures well the symmetric property of the distribution.
Looking closely at the mode quaternion of both QCQP and BNLL, 
one finds that the modes are similar for both cases.
In fact, even for the axis-symmetric case,
the mode coincides with the ``average'' quaternion $\overline{\qt}$ of sampled points.
Here $\overline{\qt}$ is defined as follows \cite{peretroukhin_so3_2020,averagingquats2007}.
\begin{equation}
    \overline{\qt}=\underset{\qt \in \Sp^3}{\arg \min } \sum_{i=1}^{N_\text{sample}} d_F\left(\qt_{i}, \qt\right)^2,
\end{equation}
where $d_F$ is the Frobenius distance defined in \eqref{eq:dist_frob}
\footnote{This is one of the Fr\'{e}chet mean. Note that using different distance gives different results.}.
%
This implies that the estimated mode quaternion heavily depends on where the sampled points come from.
If the true distribution is unimodal,
this does not a big matter since the true mode and the average of sampled quaternions are close if a large enough number of sample points $N_\text{sample}$ is taken.
If the ground truth is axis-symmetric,
however, the average of the sampled points no longer has significant information about the distribution. Thus in this case
the result of QCQP which is unimodal and whose mode is $\overline{\qt}$ fails to capture a meaningful feature of the target.

\begin{figure}[t]
    \centering
    \begin{minipage}{\linewidth}
        \centering
        \includegraphics[width=\linewidth]{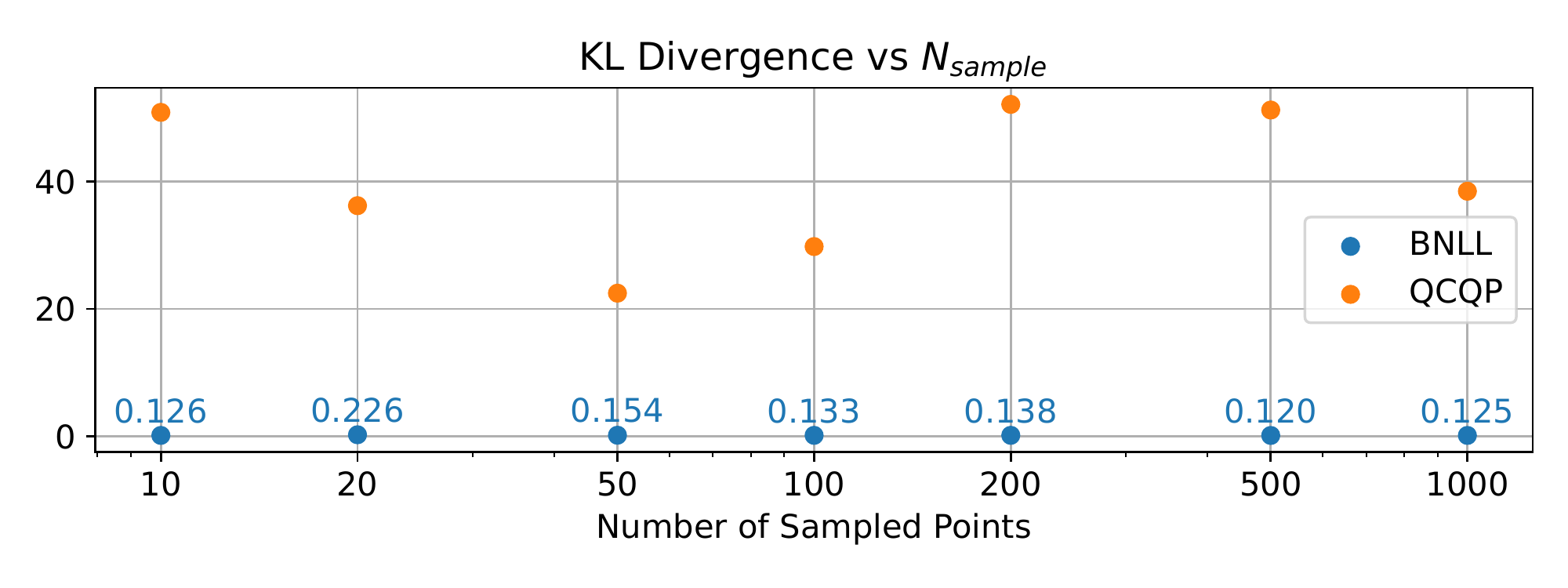}

        {\footnotesize \vspace{-3mm}(a) Variate $N_\text{sample}$, with fixed $N_\text{iter}=20000$ and $s=1.0$}
    \end{minipage}\\
    \begin{minipage}{\linewidth}
        \centering
        \includegraphics[width=\linewidth]{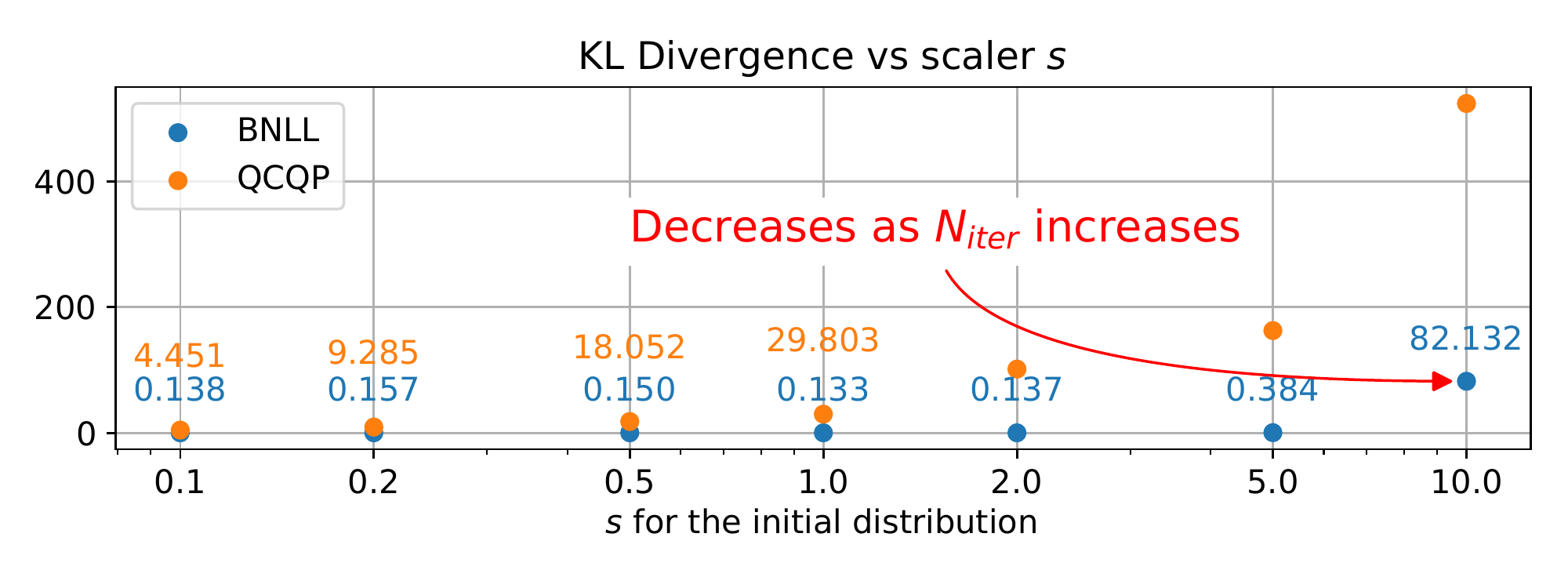}

        {\footnotesize \vspace{-3mm}(b) Variate $s$, with fixed $N_\text{iter}=20000$ and $N_\text{sample} = 100$}
    \end{minipage}
\caption{Variation of Kullback-Leibler divergence (KLD) with respect to $N_\text{sample}$ and $s$. 
At $s=10.0$ in (b), the resulting KLD of BNLL gets worse, but it decreases to 0.119 if keeping on computing until $N_\text{iter}=100000$. Note that the result of QCQP is already converged at $N_\text{iter}=20000$.}
\label{fig:KLD_vs_hyperparams}
\end{figure}

\subsection{Experiments with \textsc{ShapeNet}}


\subsubsection{Experiments Settings}

To check the performance of our loss function in the neural network, we create the evaluation network.
Here we employed a simplified PointNet structure introduced in \cite{Zhou2019}.
This network has a feature extract network $\Phi: \R^{N\times 3} \to \R^{1024}$, which consists of the sequence of 1D convolutional neural networks $3\to 64 \to 128 \to 1024$ before the max pooling layer.
$\Phi$ returns the same result for any permutation of the input points.
Given the reference points $\{P^{(i)}_{\text{ref}}\}_{i=1}^N$ and the target points $\{P^{(i)}_{\text{targ}}\}_{i=1}^N$, $z_{\text{ref}} = \Phi(P^{(1)}_{\text{ref}}, \dots, P^{(N)}_{\text{ref}})$ and $z_{\text{targ}} = \Phi(P^{(1)}_{\text{targ}}, \dots, P^{(N)}_{\text{targ}})$ are calculated. After concatenating $z_{\text{ref}}$ and $z_{\text{targ}}$, they are plugged into the MLP with dimension $2048 \to 512 \to 10$.
The 10D output is used for an inferred parameter of the distribution.

As the input data, 
we sampled point clouds from mesh models from \textsc{ShapeNet} \cite{shapenet2015}.
We used 141 models from \texttt{Airplane} category for training the network for unambiguous shape and 141 models from \texttt{Wine Bottle} category for axis-symmetric shape.
Here we call a ``unambiguous shape'' if a shape has no rotational ambiguity.
We sampled 2000 points in advance of training from each model.
In the training section, 500 points are randomly chosen from pre-sampled 2000 points at each iteration and used for a reference points $\{P^{(i)}_{\text{ref}}\}_{i=1}^N$.
A random quaternion $\qt$ is generated by normalizing a random 4-vector $\boldsymbol{v}$ sampled from normal distribution $\boldsymbol{v}\sim \mathcal{N}(0, I_4)$.
Target points $\{P^{(i)}_{\text{targ}}\}_{i=1}^N$ are generated by rotating every points in $\{P^{(i)}_{\text{ref}}\}_{i=1}^N$ with $\qt$, and a corresponding $\qt$ is used for an annotation label.

\subsubsection{Inference Results}

We inferred the rotation of the point cloud of \texttt{Airplane} with the trained network for unambiguous shape.
Similar to the result in \secref{section:sampledistr},
both QCQP and BNLL give good results for unambiguous shape.
We also examine the effect of the difference in sampled points $N_\text{sample}$.
Different sample points affect the estimation accuracy even if the number of points is the same.
In the case of $N_\text{sample}=100$, the estimation accuracy deteriorates if the points are heavily sampled from the body part.
In this case, rotational uncertainty occurs in the roll direction. Importantly, this observation is reflected in the inference result of BNLL.
It implies that the estimation accuracy would increase if the information on the roll direction is obtained.
In fact, the accuracy increases if inferring with points sampled from the wing part, which determines the roll angle of the plane.

\begin{figure*}[t]
    \centering
    \includegraphics[width=\linewidth]{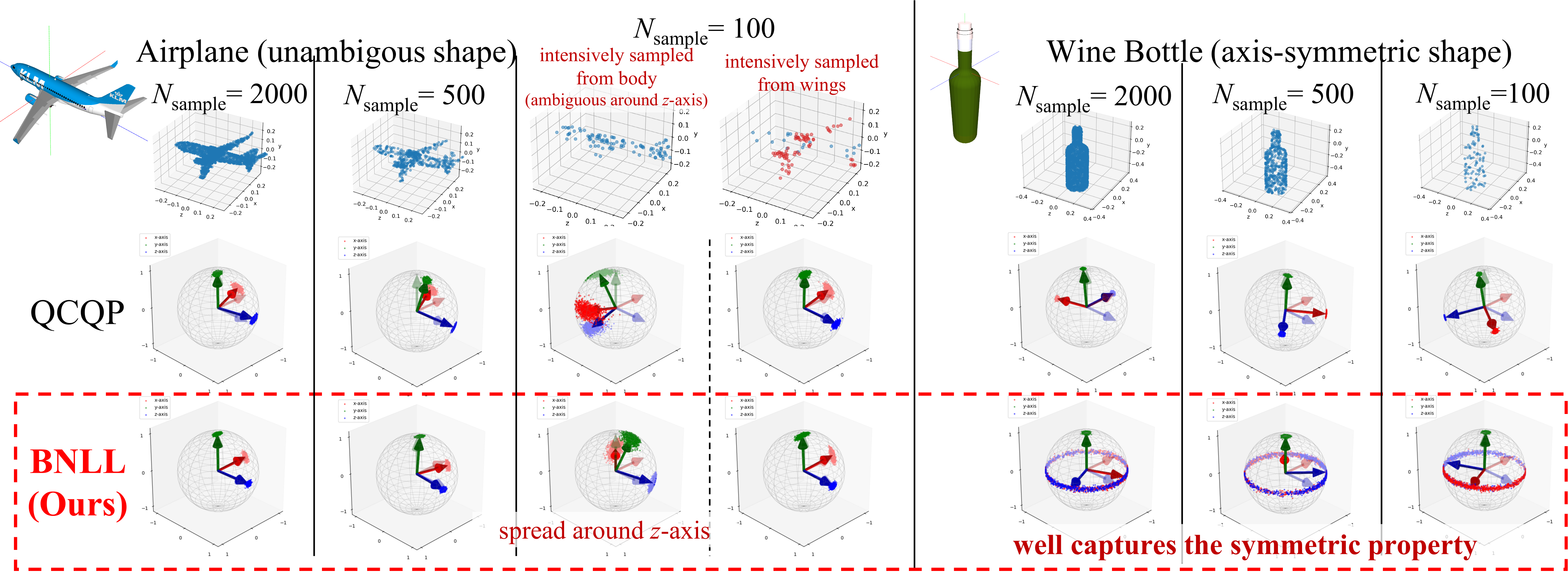}
    \caption{Examples of inference results. $N_\text{sample}$ stands for the number of points in the given point cloud.}
    \label{fig:inference_results}
\end{figure*}

In addition, 
we inferred the rotation of the point cloud of \texttt{Wine Bottle} with the trained network for axis-symmetric shape.
From \figref{fig:inference_results}, for every $N_\text{sample}=100,500,2000$, we can see that the result with BNLL can represent the axis-symmetric property very well, while QCQP gives unimodal distribution with specific rotation on the target distribution.

\section{DISCUSSIONS}

\subsubsection{Ablation Study}

Here we examine the dependency of
the number of sampled points $N_\text{sample}$ and the initial distribution $A_\text{init}$.
In this section, to remove the influence of specific object shapes such as \texttt{Wine Bottle} or \texttt{Airplane}, we tested with directly randomly generated Bingham distributions with parameter $A = D \diag(\eigvals) D^\top$.
To create $A$, we generated an orthogonal matrix $D$ with $D = \OmegaL(\qt),\, \qt \sim \bingham(O)$, and eigenvalues $\eigvals \in \R^4$ with $\eigvals \sim \mathrm{Uniform}[0, 1500)^4$. These $\eigvals$s are then shifted.

We found that the magnitude of the eigenvalues has more influence than the orthogonal matrix to be diagonalized. 
To compare without the influence of the difference in orthogonal matrices,
here we set the initial parameter given as $s\cdot A_\text{init}$ for $s \in \R$.
\figref{fig:KLD_vs_hyperparams} shows the resulting KLD after $N_\text{iter}$-th = $20000$-th iterations, with varying $N_\text{sample}$ and $s$.

\begin{figure}[t]
    \centering
    \includegraphics[width=\linewidth]{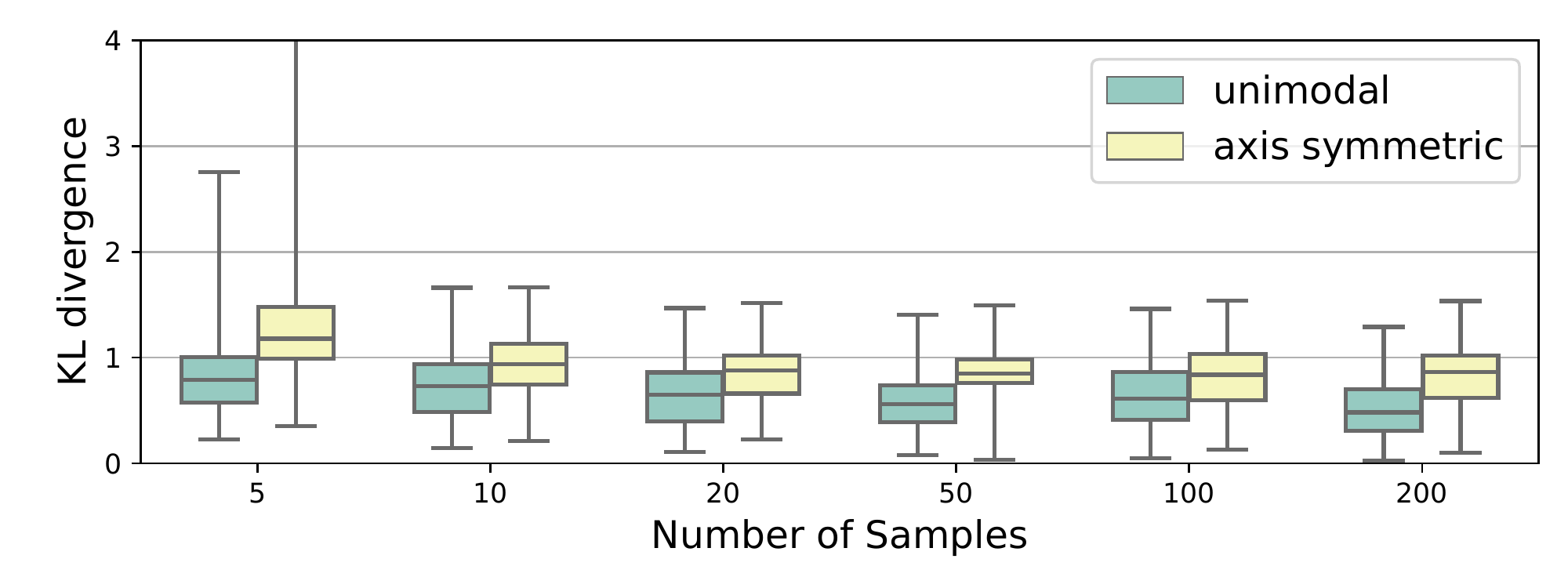}
    \caption{Resulting KLDs for some $N_\text{sample}$ after 20000 iterations. 100 times of tests were held.
    The upper and lower whiskers show the maximum and minimum values of all data (including outliers), respectively.
    In the result of axis-symmetric with $N_\text{sample} = 5$, the maximum KLD is 204.244428.
    }
    \label{fig:KLdiv_small_sample}
\end{figure}

First, we examine the effect of $N_\text{sample}$.
In general, minimizing NLL from sampled data leads to minimizing KLD for large $N_\text{sample}$ because recalling the Monte Carlo simulation,
\begin{equation}
    \int_{\qt \in \Sp^3} p(\qt) \ln(p(\qt)) \,d\qt \approx \frac{1}{N_\text{sample}} \sum^{N_\text{sample}}_{i=1} \ln(p(\qt_i))
\end{equation}
almost surely as $N_\text{sample} \to \infty$, where $\qt_i \sim \bingham(A_\text{true})$.
As shown in \figref{fig:KLD_vs_hyperparams} (a), however, we found that the resulting distribution is close enough even if $N_\text{sample}$ is quite small, say 10 or 20. 
To check if this is true when the parameter is different from $A_\text{true}$ shown in \tabref{tab:values_of_As}, 
we tested with random $A_\text{true}$ as mentioned above.
%
We optimize the distribution with the BNLL loss function 100 times.
\figref{fig:KLdiv_small_sample} shows the optimization result.
The figure implies that the KLD will be below 2.0 
even if the number of sampling points is $10$. Note that the KLD becomes very large if the sampled points are biased, and this risk may increase for small $N_\text{sample}$.



Next, we examine the dependency on an initial parameter.
\figref{fig:KLD_vs_hyperparams} (b) shows that the estimation of QCQP getting worse if the scalar $s$ increase.
The figure also implies that BNLL also slightly depends on the initial condition and ends up with a large value at $s=10$.
Even in this case, however, the KLD will decrease if we keep optimizing.
In fact, if we continue optimizing even after the 20000th iteration, KLD is still decreasing. After the 110000th iteration, it becomes 0.0905388.

If the magnitude of $\eigvals_\text{shifted}$ is too large, it may lead to slow convergence and harm the calculation's performance.
\figref{fig:KLD_vs_hyperparams} (b) implies that starting with small $\|\eigvals_\text{shifted}\|$ may lead the good convergence.
%
%
%
%
%
We empirically found that
\begin{equation}
KL(A\|O) \leq \max\{0.050, 1.5 \ln(\|\eigvals_\text{shifted}\|)\} 
\end{equation}
at least in the range $\ln(\|\eigvals_\text{shifted}\|) \leq 40$.
According to our experience,
it is fair to assume that $\ln(\|\eigvals_\text{shifted}\|) \leq 25.$
As shown in \figref{fig:transition_of_KLD}, if we start with the initial KLD around $1.5 \times 40 = 60$, it converges successfully.
As far as choosing $A=O$ for the initial distribution, BNLL seems to work well.

\subsubsection{Comparison of QCQP and BNLL}
\label{section:comparison}

By definition of QCQP loss in \eqref{eq:qcqpdef}, this only handles the mode quaternion of the parameter matrix and does not explicitly handle the eigenvalues, which determine the shape of the distribution.
Recalling our parameterization \eqref{eq:10dparam}, 
since changing the matrix elements $\bm{\theta}$ affects both the eigenvalues and eigenvectors, $\eigvals$ changes slightly as long as the mode is updated.
In contrast, 
since the BNLL contains a $\normconst(\eigvals)$ term,
it handles the eigenvalues explicitly and optimizes the distribution shape.
Therefore we can properly handle the distribution with characteristic shapes, such as axis-symmetry.

\section{CONCLUSIONS}

We proposed and implemented a Bingham NLL loss function, which is free from a pre-computed lookup table.
Our loss function is directly computable, and there is no need to interpolate computation.
We compared the performance with the QCQP loss function, a SoTA loss function of Bingham distribution that avoids the computation of normalizing constants.
We evaluated the performance of our BNLL loss function by estimating the distribution's parameter directly from sample points and by inferring a distribution using a neural network from given point clouds.
We show that our loss function can capture well the axis-symmetric property of objects.
In future works, we would like to handle mixture Bingham distribution for more capabilities, especially for the objects with discrete symmetry, based on this loss function.

\section*{ACKNOWLEDGMENT}

The visualization of Bingham distribution was supported by Ching-Hsin Fang and Duy-Nguyen Ta in Toyota Research Institute. We'd like to express our deep gratitude to them.

\addtolength{\textheight}{-10cm}   





\bibliographystyle{IEEEtran}
\bibliography{IEEEabrv,myrefs}
\end{document}